\documentclass{article} 
\usepackage{iclr2024_conference,times}
\usepackage{caption}
\usepackage{hyphenat}

\usepackage{amsmath,amsfonts,bm}

\def\eqref#1{equation~\ref{#1}}

\def\1{\bm{1}}

\DeclareMathAlphabet{\mathsfit}{\encodingdefault}{\sfdefault}{m}{sl}
\SetMathAlphabet{\mathsfit}{bold}{\encodingdefault}{\sfdefault}{bx}{n}

\newcommand{\R}{\mathbb{R}}

\usepackage{hyperref}
\usepackage{url}
\usepackage{microtype}
\usepackage{graphicx}
\usepackage{subfigure}
\usepackage{booktabs} 
\usepackage{amsthm}
\usepackage[textsize=tiny]{todonotes}
\usepackage{hyperref}
\usepackage{amssymb} 
\usepackage{multirow}
\usepackage{pbox}

\usepackage{arydshln}
\newcommand{\aucscore}[2]{#1 $\pm$ #2}
\newcommand{\bfaucscore}[2]{\textbf{#1} $\mathbf{\pm}$ \textbf{#2}}

\usepackage{tikz}

\newtheorem{theorem}{Theorem}[section]
\newtheorem{proposition}[theorem]{Proposition}

\usetikzlibrary{arrows.meta, positioning, decorations.pathreplacing, calligraphy}
\usepackage{wrapfig,lipsum,booktabs}

\usepackage{pgfplots}

\usetikzlibrary{arrows.meta, shapes.geometric, calc, positioning}

\title{\nohyphens{Rough Transformers for Continuous and Efficient Time-Series Modelling}}

\author{Fernando Moreno-Pino$^{1,}$\thanks{Equal contribution.}\hspace{1.5mm}$^{,}$\thanks{Corresponding author. Email: \texttt{fernando.moreno-pino@eng.ox.ac.uk}.}\hspace{1.75mm}, Álvaro Arroyo$^{1,*}$, Harrison Waldon$^{1,*}$, Xiaowen Dong$^{1,2}$, Álvaro Cartea$^{1,3}$  \vspace{3mm} \\
$^{1}$ Oxford-Man Institute of Quantitative Finance, University of Oxford \vspace{1mm}\\
$^{2}$ Machine Learning Research Group, University of Oxford  \vspace{1mm}\\
$^{3}$ Mathematical Institute, University of Oxford \\
}

\usepackage{mathtools}
\usepackage{amsthm,amsmath,amssymb,amsfonts,exscale,latexsym,float,eucal}
\iclrfinalcopy 
\begin{document}

\maketitle

\begin{abstract}

Time-series data in real-world medical settings typically exhibit long-range dependencies and are observed at non-uniform intervals. In such contexts, traditional sequence-based recurrent models struggle. To overcome this, researchers 
replace recurrent architectures with Neural ODE-based models to model irregularly sampled data and use Transformer-based architectures to account for long-range dependencies.
Despite the success of these two approaches, 
both incur very high computational costs for input sequences of moderate lengths and greater. To mitigate this,
we introduce the Rough Transformer, a variation of the Transformer model which operates on continuous-time representations of input sequences and incurs significantly reduced computational costs, critical for addressing long-range dependencies common in medical contexts. In particular, we propose \textit{multi-view signature attention}, which uses path signatures to augment vanilla attention and to capture both local and global dependencies in input data, while remaining robust to changes in the sequence length and sampling frequency.
We find that Rough Transformers consistently outperform their vanilla attention counterparts while obtaining the benefits of Neural ODE-based models using a fraction of the computational time and memory resources on synthetic and real-world time-series tasks. 
\end{abstract}

\vspace{-0.2cm}
\section{Introduction}\label{sec:intro}
\vspace{-0.2cm}

Real-world sequential medical data in areas such as Human Activity Recognition  \citep{chen2021deep}, suicide risk monitoring \citep{porras2022real}, or activity detection in atrial fibrillation electrograms \citep{rios2020hidden} often are irregularly sampled, of variable length, and exhibit long-range dependencies. Furthermore,
these data, which may be drawn from EEG readings \citep{vahid2020applying} or wearable devices \citep{moreno2022heterogeneous}, are often sampled at high frequency for long periods of time, yielding long sequences of data. Hence, many popular machine learning models struggle to model real-world medical sequential data, due to input dimension inflexibility, memory constraints, and computational bottlenecks.
Rather than treating these data as \textit{discrete} sequences, effective theoretical models often assume data are generated from some underlying \textit{continuous-time} process \citep{morariu2022state, ratcliff2016diffusion}. Hence, there is an increased interest in developing machine learning methods which analyze sequential medical data using \textit{continuous-time} representations. 

One recent approach to modeling continuous-time data involves the development of continuous-time analogues of standard deep learning models, such as Neural ODEs \citep{chen2018neural} and Neural CDEs \citep{kidger2020neural}, which extend ResNets \citep{he2016deep} and RNNs \citep{funahashi1993approximation}, respectively, to continuous-time settings.

Instead of processing discrete data directly, these models operate on a latent continuous-time representation of input sequences. This approach is successful in continuous-time modeling tasks where standard deep recurrent models
fail. In particular, extensions of vanilla Neural ODEs to the time-series setting \citep{rubanova2019latent, kidger2020neural} succeed
in various domains such as adaptive uncertainty quantification \citep{norcliffe2021neural}, counterfactual inference \citep{seedat2022continuous}, or generative modeling \citep{calvo2023beyond}.

In many practical medical settings, such as human behaviour changes detection  \citep{berrouiguet2018combining}, sleep activity recognition \citep{martinez2023sleep}, or heart rate fluctuations \citep{hausdorff1996multiscaled}, continuous-time data also exhibit long-range dependencies. That is, data from the distant past may impact the system's current behavior. Deep recurrent models struggle in this setting due to vanishing gradients. Several recent works \citep{melnychuk2022causal,nguyen2022transformer,moreno2023deep}  successfully extract long-range dependencies from sequential data with Transformers \citep{vaswani2017attention}, which learn temporal dependencies of a tokenized representation of input sequences. As the attention mechanism is permutation invariant, extracting such temporal dependencies requires a positional embedding of input data which projects data into some latent space.
The parallelizable nature of the Transformer allows for rapid training and evaluation on sequences of moderate length and has contributed to its success in fields such as natural language processing (NLP).

While the above approaches succeed in certain settings, several limitations hinder their wider applications. On the one hand, Neural ODEs and their analogues \citep{rubanova2019latent, kidger2020neural} bear substantial computational costs when modeling long sequences of high dimension \citep[see][]{morrill2021neural}. On the other hand, Transformers operate on discrete-time representations of input sequences, whose relative ordering is represented by the positional embedding. This representation may inhibit their expressivity in continuous-time data modeling tasks \citep{zeng2023transformers}. Moreover, Transformer-based models suffer from a number of difficulties, including (i) input sequences must be sampled at the same times, (ii) the sequence length must be fixed, and (iii) the computational cost scales quadratically in the length of the input sequence. These difficulties severely limit the application of Transformers to continuous-time data modeling.

\vspace{-0.2cm}
\section{Methods}\label{sec:model}
\vspace{-0.2cm}

\paragraph{Rough Transformer.} We propose the Rough Transformer, a Transformer-based architecture that operates on a continuous-time representation of sequential data. We make use of the the path signature from rough path theory. For a smooth function $f:\R^+ \rightarrow \R^d$, the signature $S(f)_{s, t}$ is defined as the infinite sequence of iterated integrals
\begin{align}\label{eq:signature}
    S(f)_{s, t} := \left(1, \int_s^t f'(u)\,du, \iint_{s<u_1<u_2<t} f'(u_1) \otimes f'(u_2) \,du_1 \, du_2, ...\right)\,,
\end{align}
for any $s, t \in \R^+$, where $\otimes$ denotes the tensor product; see Appendix \ref{app:sig} for details on the path signature. 

Let $\mathcal{D}$ be a dataset of irregularly sampled time-series.
To project a time-series $\mathbf{X} \in \mathcal{D}$ to a continuous-time object, we first let $\tilde{X}$ denote its piecewise-linear interpolation.
Fixing a set of times $\mathcal{T} = \{t_1, ..., t_{\overline{L}}\}$, we define the \textit{multi-view signature} for any $t_k \in \mathcal{T}$
\begin{align}
	M(\mathbf{X})_{k} := \left(S(\tilde{X})_{0, t_k}, S(\tilde{X})_{t_{k-1}, t_k}\right)
 \,.
\end{align}
In what follows, we refer to the components of $\left(S(\tilde{X})_{0, t_k}, S(\tilde{X})_{t_{k-1}, t_k}\right)$ as \textit{global} and \textit{local}, respectively. Intuitively, one can interpret the global component as an efficient representation of long-term information (see Theorem \ref{thm:universal approx} in Appendix \ref{app:sig}), and the local component as a type of convolutional filter that is invariant to the sampling rate of the signal.
Now, define the \textit{multi-view signature transform} 
\begin{align}
	M(\mathbf{X}) = (M(\mathbf{X})_{1}, ..., M(\mathbf{X})_{\bar{L}}) \,.
\end{align}
For a truncation level $n$, we define $M(\mathbf{X})^{\leq n}$ by truncating each signature component to contain terms consisting of at most $n$ integration operations. Next, define the \textit{multi-view attention mechanism}, which uses the multi-view signature transform to extend the standard attention mechanism to the space of continuous functions \citet{lyons2007differential}. First, fix a truncation level $n \in \mathbb{N}$, and let $\bar{d} \in \mathbb{N}$ be such that $M(\mathbf{X})^{\leq n}_{k} \in \mathbb{R}^{\bar{d}}$. For $h = 1, ..., H$ let $W^{\tilde{Q}, \tilde{K}, \tilde{V}}_h \in \R^{\bar{d} \times \bar{d}'} $ for some $\bar{d}' \in \mathbb{N}$. Then, let
{\footnotesize\begin{align}
    \tilde{Q}_h = M(\mathbf{X})^{\leq n}W^{\tilde{Q}}_h, \quad 
    \tilde{K}_h = M(\mathbf{X})^{\leq n}W^{\tilde{K}}_h, \quad 
    \tilde{V}_h = M(\mathbf{X})^{\leq n}W^{\tilde{V}}_h\,. 
\end{align}}
Then, the attention calculation is given by $\mathrm{softmax}\,\left(\frac{\tilde{Q}_h \tilde{K}_h^\intercal}{\sqrt{\bar{d}'}} \right)\tilde{V}_h$.
Notice that the attention calculation is similar to vanilla attention, however, we stress that the multi-view attention is built on \textit{continuous-time} objects, the signatures, while the standard attention mechanism acts on discrete objects. 


\paragraph{Advantages of Rough Transformers.} The Rough Transformer (i) is independent of the sequence length of input data, because one projects input data to its continuous-time interpolation and samples this interpolation at a fixed number of points (ii) is robust to irregular sampling, because the the iterated integrals in \eqref{eq:signature} are invariant to time-reparameterization; see Appendix \ref{app:sig} (iii) decreases the memory and computational bottleneck inherent to the vanilla Transformer, because we fix the signature sequence length to be $\overline{L} << L$. Furthermore, note that the multi-view attention mechanism does not require backpropagation through the signature calculation and can be computed \textit{in parallel}. This is significantly more computationally efficient when compared to computing signatures batch-wise in every training step.

\vspace{-0.3cm}
\section{Experiments}\label{sec:experiments}
\vspace{-0.3cm}

\paragraph{Frequency Classification.} 
Our first experiment is based on a set of synthetically generated time-series from continuous-paths of the form $\widehat{X}(t) = g(t)\sin(\omega\, t + \nu) + \eta(t)\,,$ where $g(x)$ is a non-linear trend component, $\nu$ and $\eta$ are two noise terms, and $\omega$ is the frequency. Here, the task of the model is to classify the time-series according to its frequency $\omega$. We consider $1000$ samples in $100$ classes with $\omega$ evenly distributed from $10$ to $500$. Each time-series is regularly sampled with $2000$ times-steps on the interval $[0, 6]$. This synthetic experiment is similar to others in recent work on time-series modeling \citep{li2019enhancing, yoon2019time, moreno2023deep}. To check the ability of the model to infer the true continuous-time function from irregular observations we drop $50\%$ of the data points at every epoch and report results for both the full and reduced time-series.

\begin{figure}[t!]
\centering
	\includegraphics[width=0.4\linewidth]{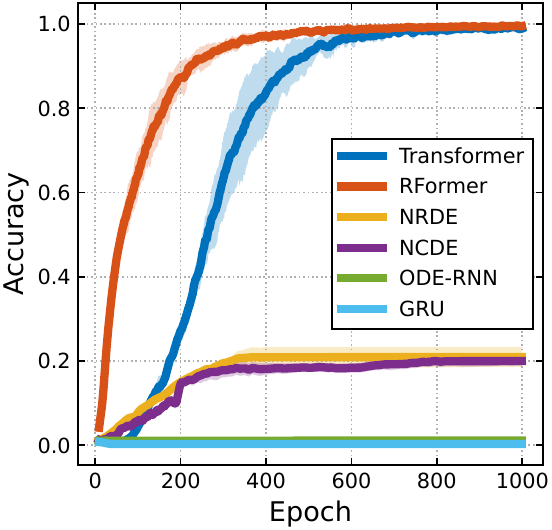}
    \includegraphics[width=0.4\linewidth]{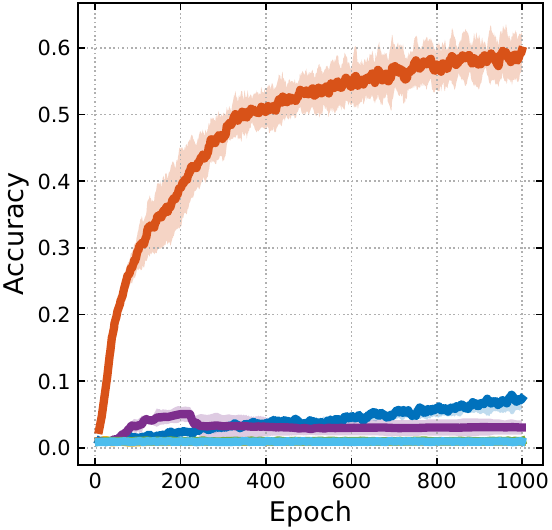}
    \vspace{-0.2cm}

	\caption{Test accuracy per epoch for the frequency classification task across three random seeds. \textbf{Left:} Performance for the full time-series. \textbf{Right:} Performance when randomly dropping half of the datapoints every epoch.}
	\label{fig:test_full_synthetic}
\end{figure}

Figure \ref{fig:test_full_synthetic} shows that the Rough Transformer, hereafter denoted \texttt{RFormer}, improves the performance of the Transformer in two ways. First, the inclusion of both local and global information through the multi-view signature enhances the sample efficiency of the model, even though the attention mechanism is now operating on a much shorter sequence. Second, the model is robust to changes in the sampling frequency at inference time, despite operating with half of the context. Therefore, in this task, unlike the vanilla Transformer, \texttt{RFormer} adapts to changes in both the length and the sampling frequency of the input stream.\footnote{We highlight that the drop in performance is likely due to aliasing, i.e., the sampling rate is below the Nyquist rate, meaning that higher frequency signals cannot be appropriately represented with the number of data points available.}

When compared with other models, we see that \texttt{GRU} and \texttt{ODE-RNN} fail to capture the information in the signal, and are not able to obtain any meaningful performance improvement throughout the training period. This highlights the shortcomings of most RNN-based models when processing sequences of moderate length, which are very common in real-world applications. In turn,  \texttt{Neural-CDE} and \texttt{Neural-RDE} \citep{morrill2021neural}, which is tailor-made for long-range time series modeling, can capture some useful dependencies in the time series but falls short when compared with both vanilla \texttt{Transformer} and \texttt{RFormer}. When data are irregularly sampled, we see a very sharp drop in performance for the \texttt{Transformer} model. Both \texttt{Neural-CDE} and \texttt{Neural-RDE} also exhibit a significant drop in performance despite being continuous-time.\footnote{In the case of NRDEs, this phenomenon was observed consistently for both shorter and longer windows.}

\vspace{-0.3cm}

\paragraph{HR dataset.} Next, we consider the Heart Rate dataset from the TSR archive \citep{tan2020monash}, originally sourced from Beth Israel Deaconess Medical Center (BIDMC). This dataset consists of time-series sampled from patient ECG readings, and each model is tasked with forecasting the patient's heart rate (HR) at the sample's conclusion. 
The data, sampled at 125Hz, consists of three-channel time-series (including time), each spanning $4000$ time steps.
We used the L2 loss metric to assess the performance. Table \ref{table:hr} shows the results.

The sequences in the HR dataset are sufficiently short to remain within memory when running the \texttt{Transformer} model. The baseline \texttt{Transformer} model improves over \texttt{GRU} and \texttt{ODE-RNN}, however it is less competitive when compared with \texttt{Neural-RDE}, suggesting that the Transformer is not particularly well-suited for this type of task. However, the \texttt{RFormer} model improves over the baseline \texttt{Transformer} by $171\%$ and yields the best performance overall, experiencing a very small fall in test loss when it comes to the random drop, similar to other continuous-time models.

\begin{table}[t!]
	\centering
	\caption{Test RMSE (mean $\pm$ std), computed across five seeds on the Heart Rate (HR) dataset. The results correspond to both our reproduction and those presented in \citep{morrill2021neural}. }\label{table:hr}
	\begin{center}
		\begin{small}
			\begin{tabular}{@{}lcc@{}}
				\toprule
				\multirow{2}{*}{Model} & \multicolumn{2}{c}{Test RMSE}  \\ \cmidrule(l){2-3} 
				& Full & 50\% Drop \\ \midrule       
				ODE-RNN$^\diamond$ & \aucscore{13.06}{0.00} & - \\
                Neural CDE$^\diamond$ & \aucscore{9.82}{0.34} &  - \\
                Neural RDE$^\diamond$ & \aucscore{2.97}{0.45} & - \\
                [0.1ex] \hdashline\noalign{\vskip 0.7ex}
                GRU$^\dagger$ & \aucscore{13.06}{0.00} &  \aucscore{13.06}{0.00}  \\
                ODE-RNN$^\dagger$ & \aucscore{13.06}{0.00} & \aucscore{13.06}{0.00} \\ \noalign{\vskip 0.7ex}
                Neural RDE$^\dagger$& \aucscore{\underline{4.04}}{0.11} & \aucscore{\underline{4.67}}{0.40} \\ \noalign{\vskip 0.7ex}
				Transformer & \aucscore{8.24}{2.24} & \aucscore{21.01}{3.81}  \\[0.1ex] \hdashline\noalign{\vskip 0.7ex}
				\textbf{RFormer}  & \bfaucscore{3.04}{0.03} & \bfaucscore{3.31}{0.05}   \\ \bottomrule
			\end{tabular}
		\end{small}
	\end{center}
\end{table}

\vspace{-0.1cm}

\paragraph{Training Efficiency.} Attention-based architectures are highly parallelizable on modern GPUs, as opposed to traditional RNN models which require sequential updating. However, vanilla attention experiences a bottleneck in memory and time complexity as the sequence length $L$ grows.
As covered above in Section \ref{sec:model}, variations of the signature transform allow the model to operate on a reduced sequence length $\bar{L}$ without increasing the dimensionality in a way that would become problematic for the model. This allows us to bypass the quadratic complexity of the model without resorting to sparsity techniques commonly used in the literature \citep{li2019enhancing, feng2023diffuser}. The previous experiment demonstrates empirically that operating on this continuous-time representation does not hinder (and in fact improves) model performance, with the added benefit of robustness to irregularly-sampled time-series. Table \ref{table:times} shows the empirical running times for the optimal signature hyperparameters (which can be found in Appendix \ref{app:exp_details}), with all models run on the same machine and using the same batch size. We also report the speedup of the \texttt{RFormer} model when compared to the vanilla \texttt{Transformer}.

\begin{table}[]
	\centering
    	\caption{Seconds per epoch for all models considered. See Appendices \ref{app:hyper} and \ref{app:exp_details} for experimental details.}
     \label{table:times}
	\begin{center}
		\begin{footnotesize}
			\begin{tabular}{@{}lcc@{}}
				\toprule
				\multirow{2}{*}{Model} & \multicolumn{2}{c}{Running Time (seconds / epoch)}  \\ \cmidrule(l){2-3} 
				& Sine & HR\\ \midrule
                GRU & \textbf{0.12} &\underline{1.07} \\
				ODE-RNN & 5.39 &  50.71\\
                    Neural CDE & 9.83 & - \\
				Neural RDE & 0.85 & 9.52 \\
				Transformer & 0.77 & 11.71 \\[0.1ex] \hdashline\noalign{\vskip 0.7ex}
				\textbf{RFormer}  & \underline{0.55} & \textbf{0.45 }\\
                    \textbf{Speedup} & $1.4\times$ & $26.11\times$  \\
                \bottomrule
			\end{tabular}
		\end{footnotesize}
	\end{center}
\end{table}

The computational efficiency gains of \texttt{RFormer} are attained due to the multi-view signature transform reducing the length of the time-series with minimal information loss. 
This contrasts with NRDEs, which augment NCDEs 
with local signatures of input data, and find that smaller windows often perform better. Hence, NRDEs do not experience the same computational gains as \texttt{RFormer}, as they must perform many costly ODE integration steps.

\vspace{-0.2cm}
\section{Conclusion}
\vspace{-0.2cm}

In this paper, we introduced the Rough Transformer, a variant of the original Transformer that allows the processing of discrete-time series as continuous-time signals through the use of multi-view signature attention. Empirical comparisons showed that Rough Transformers outperform vanilla Transformers and continuous-time models on a variety of time-series tasks and are robust to the sampling rate of the signal. Finally, we show that RFormer provides significant speedups in training time compared to regular attention and ODE-based methods, without the need for major architectural modifications or sparsity constraints. In future work, it would be interesting to see if Rough Transformers also provide good performance in other application domains such as finance \citep{scalzo2021nonstationary,moreno2022deepvol,arroyo2022dynamic, arroyo2024deep} or weather forecasting \citep{nguyen2023climax}.\\

\bibliography{iclr2024_conference}
\bibliographystyle{iclr2024_conference}

\newpage
\appendix

\newpage

\section{Properties of Path Signatures}
\label{app:sig}

First, we recall that smooth paths are uniquely determined by their signatures, which motivates its use as a feature map. 
\begin{proposition}
    Given a smooth path $\widehat{X}: [0, T] \rightarrow \R^d$, then the map $P_{\widehat{X}}: [0, T] \rightarrow \R^{1+d}$ where $P_{\widehat{X}}(t)= (t, \widehat{X}(t))$ is uniquely determined by it's signature $S(P_{\widehat{X}})_{0, T}$.
\end{proposition} 
The proof of this fact in \citep{hambly2010uniqueness}. 

For Rough Transformers, several features of path signatures are important. First, linear functionals on path signatures possess universal approximation properties for continuous functionals.
\begin{theorem}\label{thm:universal approx}
	Fix $T>0$, and let $K \subset C^1_b([0, T]; \R^d)$. Let $f : C^1_b([0, T]; \R^{d+1}) \rightarrow \R$ be continuous with respect to the sup-norm topology on $C^1_b([0, T]; \R^{d+1})$. Then for any $\epsilon>0$, there exists a linear functional $\ell$ such that
	\begin{align}
		| f(P_{\widehat{X}}) - \langle \ell, S(P_{\widehat{X}})_{0, T}\rangle| \leq \epsilon\,,
	\end{align}
	for any $\widehat{X}\in K$. 
\end{theorem}
For a proof of \ref{thm:universal approx}, see \citep{arribas2018derivatives}. Even though Theorem \eqref{thm:universal approx} guarantees that \textit{linear} functionals are sufficient for universal approximation, linear models are not always sufficient in practice. This motivates the development of nonlinear models built upon the path signature which efficiently extract path behavior.

The second feature is that the terms of the path signature decay factorially, as described by the following proposition.
\begin{proposition}\label{prop:decay}
	Given $\widehat{X} \in C^1_b([0, T]; \R^d)$, for any $s, t \in [0, T]$, we have that for any $I \in \mathcal{I}^n_d$
	\begin{align}
		|S(\widehat{X})_{0, T}^I| = O\left(1/n!\right)\,.
	\end{align}
\end{proposition}
For a proof of Proposition \ref{prop:decay}, see \citep{lyons2007differential}. Hence, the number of terms in signature grows exponentially in the level of the signature, but the tail of the signature is well-behaved, so only a few levels in a truncated signature are necessary to adequately approximate continuous functionals.

\subsection{Signatures of Piecewise Linear Paths.}
In the Rough Transformer, we use linear interpolation of input time-series to get a continuous-time representation of the data. As mentioned in Section \ref{sec:model}, the signature computation in this case is particularly simple. 

Suppose $\widehat{X}_k: [t_k, t_{k+1}] \rightarrow \R^d$ is a linear interpolation between two points $X_k, X_{k+1} \in \R^d$. That is,
\begin{align}
    \widehat{X}_k(t) = X_{k} + \frac{t - t_k}{t_{k+1} - t_k}\left(X_{k+1} - X_k\right)\,.
\end{align}
Then the signature of $\widehat{X}_k$ is given explicitly:
\begin{align}
    S(\widehat{X}_k)_{t_k, t_{k+1}} = \left(1, X_{k+1} - X_k, \frac{1}{2} (X_{k+1} - X_k)^{\otimes 2}, \frac{1}{3!}(X_{k+1} - X_k)^{\otimes 3}, ..., \frac{1}{n!}(X_{k+1} - X_k)^{\otimes n}, ...\right)\,,
\end{align}
where $\otimes$ denotes the tensor product. Let $\widehat{X}_k * \widehat{X}_{k+1}$ denote the \textit{concatenation} of $\widehat{X}_k$ and $\widehat{X}_{k+1}$. That is, $\widehat{X}_k * \widehat{X}_{k+1} : [t_k, t_{k+2}] \rightarrow \R^d$ is defined
\begin{align}
    \widehat{X}_k * \widehat{X}_{k+1}(t) = \begin{cases}
        \widehat{X}_k(t) & t\in [t_k, t_{k+1}] \\
        \widehat{X}_{k+1}(t) & t\in (t_2, t_{k+2}]\,.
    \end{cases}
\end{align}
The signature of the concatenation $\widehat{X}_k * \widehat{X}_{k+1}$ is given by \textit{Chen's relation}.
\begin{proposition}[Chen's Relation]
The following identity holds:
    \begin{align}
        S(\widehat{X}_k * \widehat{X}_{k+1})_{t_{k}, t_{k+2}} = S(\widehat{X}_k)_{t_{k}, t_{k+1}} \otimes S(\widehat{X}_{k+1})_{t_{k+1}, t_{k+2}}\,,
    \end{align}
    where for elements $A, B \in T((\R^d))$ with $A = (A_0, A_1, A_2, ... )$ and $B = (B_0, B_1, B_2, ...)$ the tensor product $\otimes$ is defined
    \begin{align}
        A \otimes B = \left(\sum_{j = 0}^k A_j \otimes B_{k - j}\right)_{k\geq0}\,.
    \end{align}
\end{proposition}
Let $\mathbf{X} = (X_0, ..., X_L)$ be a time-series. Then the linear interpolation $\tilde{X}:[0, T] \rightarrow \R^d$ can be represented as the concatenation of a finite number of linear paths:
\begin{align}
    \tilde{X} = \widehat{X}_0 * \cdots * \widehat{X}_{L-1}\,.
\end{align}
Hence, the signature can be computed:
\begin{align}
    S(\tilde{X})_{0, T} = S(\widehat{X}_0)_{0, t_1} \otimes \cdots \otimes S(\widehat{X}_{L-1})_{t_{L-1}, T}\,.
\end{align}

\newpage

\section{Experimental Details}
\label{app:exp_details}
All experiments were conducted on an NVIDIA GeForce RTX 3090 GPU with 24,564 MiB of memory, utilizing CUDA version 12.3. The timings presented in Table \ref{table:times} were obtained by executing each model independently for each dataset and averaging the resulting times across 100 epochs. We take down the hyperparameters used in the \texttt{RFormer} model for each of the datasets in the paper, which were chosen in accordance with the performance of the model in the validation set.

\begin{table}[H]
\centering
	\caption{Validation accuracy on the sinusoidal dataset.}\label{table:nRDE_validation_sin}
 \begin{center}
 \noindent
  \makebox[\textwidth]{
\begin{tabular}{cccccc}
\toprule
\textbf{Dataset}        & \textbf{Learning Rate} & \textbf{Number of Windows} & \textbf{Sig. Depth} & \textbf{Sig. Type}  \\ \toprule
Sinusoidal & $1\times10^{-3}$ & 75 &  2 & Multi-View       \\
HR & $1\times10^{-3}$ & 75 & 4 & Local      \\
 \toprule 
\end{tabular}}
\end{center}
\end{table}

\newpage

\section{Models Validation}
\label{app:hyper}

This section collects the validation of Step and Depth for the neural RDE model in each of the evaluated datasets. Optimal values are selected for evaluation on test-set. Early-stopping was used with same criteria than \citealp{morrill2021neural}.

\begin{table}[H]
\centering
	\caption{Validation accuracy on the sinusoidal dataset.}\label{table:nRDE_validation_sin}
 \begin{center}
\begin{tabular}{ccccc}
\toprule
\textbf{Acc. Val}        & \textbf{Step} & \textbf{Depth} & \textbf{Memory Usage (Mb)} & \textbf{Elapsed Time (s)} \\ \toprule
17.26          & 2             & 2              & 778.9                  & 6912.7                 \\
12.21          & 2             & 3              & 770.3                  & 1194.43                \\
16.35          & 4             & 2              & 382.2                  & 2702.48                \\
19.27          & 4             & 3              & 386.16                 & 574.97                 \\
20.99          & 8             & 2              & 193                    & 1321.36                \\
\textbf{24.02} & \textbf{8}    & \textbf{3}     & \textbf{194.17}        & \textbf{332.17}        \\
17.15          & 16            & 2              & 97.13                  & 136.43                 \\
21.59          & 16            & 3              & 98.17                  & 156.93                 \\
17.46          & 24            & 2              & 65.96                  & 105.94                 \\
20.59          & 24            & 3              & 66.68                  & 98.97             \\    \toprule 
\end{tabular}
\end{center}
\end{table}

\begin{table}[H]
\centering
	\caption{Validation loss on the HR dataset.}\label{table:nRDE_validation_HR}
 \begin{center}
\begin{tabular}{ccccc}
\toprule \textbf{Acc. Val} & \textbf{Step} & \textbf{Depth} & \textbf{Memory Usage (Mb)} & \textbf{Elapsed Time (s)} \\ \toprule
\textbf{2.44}     & \textbf{2}    & \textbf{2}     & \textbf{5044.44}       & \textbf{56492.33}      \\
3.03              & 2             & 3              & 5059.28                & 39855.19               \\
3.67              & 4             & 2              & 2515.40                & 10765.58               \\
16.04             & 4             & 3              & 2531.44                & 7157.20                \\
5.35              & 8             & 2              & 1259.30                & 3723.94                \\
2.70              & 8             & 3              & 1268.60                & 18682.82               \\
3.58              & 16            & 2              & 632.08                 & 3518.96                \\
3.64              & 16            & 3              & 636.64                 & 7922.96                \\
3.86              & 24            & 2              & 422.74                 & 3710.95                \\
3.55              & 24            & 3              & 426.83                 & 6567.02  \\ \bottomrule             
\end{tabular}
\end{center}
\end{table}

\newpage

\newpage

\end{document}